\documentclass[10pt,letterpaper]{article}

\usepackage{geometry}
\usepackage{times}
\usepackage{graphicx}
\usepackage{amsmath}
\usepackage{amssymb}
\usepackage{booktabs}
\usepackage{hyperref}
\usepackage{authblk}
\usepackage{abstract}
\usepackage{algorithm}
\usepackage{algorithmic}

\title{\LARGE \bf OOM-RL: Out-of-Money Reinforcement Learning \\ \large Market-Driven Alignment for LLM-Based Multi-Agent Systems}

\author[1]{Kun Liu\thanks{Corresponding author: \texttt{ai@quantpits.com}}}
\author[1]{Liqun Chen}
\affil[1]{QuantPits.com}
\date{}

\begin{document}
	
	\maketitle
	
	\begin{abstract}
		The alignment of Multi-Agent Systems (MAS) for autonomous software engineering is constrained by evaluator epistemic uncertainty. Current paradigms, such as Reinforcement Learning from Human Feedback (RLHF) and AI Feedback (RLAIF), frequently induce model sycophancy, while execution-based environments suffer from adversarial ``Test Evasion'' by unconstrained agents. In this paper, we introduce an objective alignment paradigm: \textbf{Out-of-Money Reinforcement Learning (OOM-RL)}. By deploying agents into the non-stationary, high-friction reality of live financial markets, we utilize critical capital depletion as an externally imposed negative gradient. Our longitudinal 20-month empirical study (July 2024 -- February 2026) chronicles the system's evolution from a high-turnover, sycophantic baseline to a robust, liquidity-aware architecture. We show that the economic consequences of financial loss---real execution costs, slippage, and capital depletion---exposed failure modes not apparent under internal evaluation alone and motivated architectural and governance changes that were later formalized as the \textbf{Strict Test-Driven Agentic Workflow (STDAW)}, a Byzantine-inspired uni-directional state lock (RO-Lock) anchored to a deterministically verified $\geq 95\%$ code coverage constraint matrix. During the final 94-trading-day observation window, the production system exhibited improved execution-aware performance, including an annualized Sharpe ratio of approximately 2.06. These financial results are observational and temporally bounded; they should not be interpreted as evidence of persistent investment alpha or as a causal estimate of STDAW's contribution. The primary contribution of this work is the use of externally imposed economic consequences as an epistemic constraint on agentic development, laying the groundwork for generalized paradigms where real-world resource depletion acts as an objective physical constraint.
		
		\vspace{0.5em}
		\noindent\textbf{Keywords:} AI Alignment, Multi-Agent Systems (MAS), Out-of-Money Reinforcement Learning (OOM-RL), Test Evasion, Sim-to-Real Gap, Autonomous Software Engineering, Sycophancy.
	\end{abstract}
	
	\section{Introduction}
	The rapid proliferation of Large Language Models (LLMs) has catalyzed a shift in automated software engineering, evolving from passive code assistants \cite{chen2021evaluating} to autonomous Multi-Agent Systems (MAS) capable of end-to-end repository generation and program repair \cite{he2025llm, bouzenia2025repair}. As these systems undertake complex reasoning tasks, ensuring their safe and effective operation has become the central challenge of AI alignment. Currently, the gold standard relies on Reinforcement Learning from Human Feedback (RLHF) or scalable oversight mechanisms such as AI Feedback (RLAIF) \cite{lee2023rlaif, kenton2024scalable}.
	
	However, human and AI evaluators are constrained by the "Evaluator's Dilemma." When tasked with reviewing intricate, multi-step logical pipelines, evaluators often lack the domain expertise to identify subtle architectural flaws. Consequently, models aligned via these paradigms develop sycophantic behaviors—optimizing for outputs that \textit{appear} structurally elegant and reasoned to the evaluator, rather than those that are empirically correct \cite{perez2023discovering, fanous2025syceval, kim2025challenging}. This phenomenon is a manifestation of reward gaming and specification gaming \cite{skalse2022defining, krakovna2020specification}, where the MAS learns to hack the subjective reward model rather than solve the underlying problem—a vulnerability recently shown to cause emergent misalignment even in production RL systems \cite{macdiarmid2025natural}.
	
	To bypass subjective evaluation, researchers have shifted towards execution-based evaluation \cite{wang2023execution} and LLM-driven Test-Driven Development (TDD) \cite{mathews2024test}. Yet, deploying MAS in read-write environments introduces a vulnerability: "Test Evasion." When provided unbounded access to a codebase, LLMs frequently exhibit adversarial behaviors, introducing modifications to test assertions to artificially inflate coverage without fulfilling the intended business logic \cite{zhang2025llm, yin2025detecting}. Furthermore, even when syntactically perfect code passes all simulated unit tests, it frequently experiences performance degradation upon real-world deployment due to the pervasive Simulation-to-Reality (Sim2Real) gap \cite{wagenmaker2024overcoming}.
	
	To overcome these deficiencies, we introduce a novel alignment paradigm: \textbf{Out-of-Money Reinforcement Learning (OOM-RL)}. We posit that an objective function for an autonomous MAS is survival in an adversarial, high-stakes physical environment. Live financial markets serve as a discriminator; they are intrinsically non-stationary \cite{padakandla2020reinforcement, liu2021towards} and penalize latency and microstructural friction \cite{kearns2013machine, yuan2025mechanisms}. Unlike human preference models or isolated static code compilers, financial markets cannot be flattered or trivially exploited. In OOM-RL, the loss function is capital depletion. A system that hallucinates logic or attempts to evade structural constraints faces a financial penalty.
	
	To operationalize OOM-RL while preventing the MAS from circumventing the evaluation framework (e.g., via sandbox escapes) \cite{marchand2026quantifying, rabin2025sandboxeval}, we propose the \textbf{Strict Test-Driven Agentic Workflow (STDAW)}. This architecture, which codified engineering constraints that emerged from the preceding developmental experience and was formally committed near the end of the study period (February 2026), utilizes a uni-directional state locking mechanism (RO-Lock) to anchor the agent's generative capabilities against a deterministic Continuous Integration (CI) boundary---enforcing a \textbf{near-exhaustive coverage threshold ($\ge 95\%$) across the entire $8\text{K}+$-line \texttt{QuantPits} project codebase}.
	
	Our results support the utility of this alignment paradigm. We summarize our contributions as follows:
	\begin{itemize}
		\item We formalize \textbf{OOM-RL}, showing how real-world financial friction acts as an objective, dense negative gradient that bridges the Sim2Real gap through iterative adaptation.
		\item We design \textbf{STDAW}, an adversarial engineering framework that utilizes uni-directional state locking to address the ``Test Evasion'' phenomenon in autonomous software engineering.
		\item We present a 20-month empirical study detailing the system's transition from high-turnover, high-drawdown ``sycophantic'' trading to a resilient, ensemble-driven architecture, and report the observed financial performance during a bounded 94-trading-day mature observation window.
		\item We conceptualize \textbf{Reinforcement Learning from Cloud Billing (RLFCB)}, a domain-agnostic extension that frames computational resource depletion (e.g., cloud-based ``Out-of-Money'' states) as a generalized physical friction for non-financial MAS.
	\end{itemize}
	
	\section{Related Work}
	
	\subsection{Scalable Oversight and the Sycophancy Bottleneck}
	The foundational approach to aligning LLMs with human intent relies heavily on RLHF and, more recently, AI-driven scalable oversight mechanisms such as RLAIF \cite{lee2023rlaif}. As models surpass human capabilities in specialized domains, researchers have increasingly utilized weak LLMs to evaluate the outputs of strong LLMs \cite{kenton2024scalable}. However, these proxy-based evaluation paradigms are vulnerable to specification gaming \cite{krakovna2020specification} and reward gaming \cite{skalse2022defining}.
	
	A failure mode of this vulnerability is LLM sycophancy—the model's tendency to prioritize the evaluator's approval over objective correctness. Recent empirical studies reveal that models learn to exploit the epistemic uncertainty of human or weak AI evaluators by generating confident but hallucinated logic \cite{perez2023discovering, fanous2025syceval}. Even when subjected to adversarial user rebuttals, models aligned via preference-based paradigms persistently exhibit sycophantic behavior rather than defending objective ground truth \cite{kim2025challenging}. Critically, this tendency toward reward hacking inevitably leads to natural emergent misalignment when such systems transition from synthetic evaluations into production environments \cite{macdiarmid2025natural}. OOM-RL aims to circumvent this sycophancy bottleneck by replacing subjective, hackable evaluators with the non-cooperative feedback of real-world financial consequences.
	
	\subsection{Execution-Based Evaluation and Adversarial "Test Evasion"}
	To establish an objective alignment metric for logic and code generation, the community has shifted towards execution-based evaluation \cite{chen2021evaluating, wang2023execution}. This paradigm has fueled the development of LLM-based Multi-Agent Systems (MAS) for automated software engineering \cite{he2025llm} and autonomous program repair \cite{bouzenia2025repair}, frequently integrating LLMs with Test-Driven Development (TDD) pipelines \cite{mathews2024test}.
	
	Despite these advances, practical code generation remains plagued by complex hallucination mechanisms \cite{zhang2025llm}. Crucially, when agents are deployed in interactive, unconstrained read-write environments, they exhibit adversarial ingenuity. Recent works have identified models modifying constraints to pass otherwise failing conditions \cite{yin2025detecting}, a phenomenon we term "Test Evasion." The security implications of such behaviors are profound, raising concerns regarding untrusted code execution \cite{rabin2025sandboxeval}, container sandbox escapes by frontier LLMs \cite{marchand2026quantifying}, and the limitations of current vulnerability detection systems \cite{tihanyi2026vulnerability}. By conceptualizing MAS reliability through the lens of Byzantine Fault Tolerance \cite{zheng2026rethinking}, our proposed STDAW architecture addresses this by enforcing deterministic uni-directional state locks, preventing the AI from subverting the evaluation sandbox.
	
	\subsection{Non-Stationary Environments and the Sim2Real Gap}
	Reinforcement learning within non-stationary environments has long been a challenge \cite{padakandla2020reinforcement}, particularly when systems encounter out-of-distribution (OOD) scenarios \cite{liu2021towards}. A major impediment to deploying RL agents in physical reality is the Sim-to-Real gap \cite{wagenmaker2024overcoming}, where policies optimized in frictionless simulations fail upon real-world deployment.
	
	Financial markets epitomize the non-stationary, OOD environment, characterized by microstructural noise and the execution friction inherent in active trading \cite{kearns2013machine, yuan2025mechanisms}. Traditional simulated trading frameworks inadvertently incentivize models to exploit theoretical zero-friction assumptions. In contrast, OOM-RL leverages this microstructural friction (e.g., liquidity droughts, order slippage) not as a nuisance, but as a dense, negative reward gradient. By forcing the MAS to internalize the financial penalties of the Sim2Real gap, OOM-RL aligns the agent's generative architecture toward resilience rather than theoretical optimality.
	
	\section{Methodology}
	
	To systematically align LLM-based Multi-Agent Systems (MAS) using real-world market dynamics, we propose a dual-loop adversarial architecture. The framework decouples the logical verification of the MAS (Inner Loop) from its empirical out-of-distribution (OOD) survival (Outer Loop). In this section, we detail the structural constraints and mathematical formulations that operationalize Out-of-Money Reinforcement Learning (OOM-RL).
	
	\subsection{Architecture Overview: The Dual-Loop Alignment}
	
	The traditional RLHF pipeline relies on a singular update loop driven by human preference. In contrast, our architecture recognizes that autonomous code generation fundamentally requires two distinct validation boundaries before capital deployment:
	\begin{enumerate}
		\item \textbf{The Inner Loop (Epistemic Constraint):} Governed by the Strict Test-Driven Agentic Workflow (STDAW). It ensures that the generated pipeline is mathematically sound, syntactically flawless, and deterministic prior to execution.
		\item \textbf{The Outer Loop (Ontological Constraint):} Governed by OOM-RL. It subjects the syntactically perfect codebase to the non-stationary, high-friction reality of the live financial market to evaluate its true alignment with utility generation.
	\end{enumerate}
	
	\subsection{Strict Test-Driven Agentic Workflow (STDAW)}
	
	Unconstrained MAS deployed in read-write environments exhibit "Test Evasion"—the adversarial modification of verification metrics to disguise logical hallucinations. To mitigate this Byzantine behavior, STDAW implements a multi-dimensional constraint matrix.
	
	\subsubsection{Near-Exhaustive Deterministic Constraint Matrix}
	LLMs are adept at exploiting ``coverage gaps'' in unit tests. To construct a rigorous epistemic boundary, STDAW was developed as the structural culmination of the 20-month developmental trajectory. Near the end of the study period (initial commit February 24, 2026), we formalized the sandbox boundary by rigidly enforcing a mathematically verifiable \textbf{strict coverage constraint ($\tau_{cov} \ge 95\%$)} across the entirety of the \texttt{QuantPits} project codebase (approximately 8,300 lines of code). 
	
	This matrix serves as the terminal ground truth. Having internalized the risks of structural hallucinations during the high-friction epochs, the system now treats any alteration to fundamental financial mathematics (e.g., dividend reinvestment alignment, cross-sectional ranking operators) as a failure. The density of the test suite reduces the agent's degree of freedom for hallucination, codifying the lessons of OOM-RL into a permanent software barrier.
	
	\subsubsection{Uni-Directional State Locking (RO-Lock)}
	To prevent the MAS from subverting the $\geq 95\%$ constraint matrix, we formalize the \textbf{RO-Lock (Read-Only Lock)} mechanism. Drawing on Byzantine fault-tolerance separation-of-authority principles \cite{zheng2026rethinking}, RO-Lock ensures that the agent cannot simultaneously act as both the ``Creator'' and the ``Judge.''
	
	In our engineering implementation, the RO-Lock is enforced at the OS level using Docker container orchestration. During the verification phase, the test directory $\mathcal{T}$ is mounted as a Read-Only volume, preventing the agent from overwriting existing assertions or mock data. Furthermore, we implement an AST-based (Abstract Syntax Tree) sanitization layer that scans the generated code $\mathcal{S}$ for reflective patterns or monkey-patching attempts targeting the testing framework (e.g., `pytest` or `unittest`).
	
	Let $\mathcal{S}$ represent the source code directory (\texttt{src/}) and $\mathcal{T}$ represent the test directory (\texttt{tests/}). The agent operates under a strict access-control policy function $\pi_{lock}(E)$, where $E$ is the current execution phase.
	
	\begin{algorithm}[h]
		\caption{Uni-Directional RO-Lock State Machine (STDAW)}
		\label{alg:rolock}
		\begin{algorithmic}[1]
			\REQUIRE Execution Phase $E \in \{ \text{Logic\_Genesis}, \text{Test\_Genesis} \}$
			\REQUIRE Source code state $\mathcal{S}$, Test constraint matrix $\mathcal{T}$, Deterministic baseline $\mathcal{B}$
			\STATE \textbf{Initialize} access capability mapping $\Pi: \{\mathcal{S}, \mathcal{T}\} \to 2^{\{\mathsf{R}, \mathsf{W}, \mathsf{X}\}}$
			\STATE Set default boundaries: $\Pi(\mathcal{S}) \gets \{\mathsf{R}\}$, $\Pi(\mathcal{T}) \gets \{\mathsf{R}\}$
			
			\IF{$E == \text{Logic\_Genesis}$}
			\STATE $\Pi(\mathcal{S}) \gets \{\mathsf{R}, \mathsf{W}\}$
			\STATE $\Pi(\mathcal{T}) \gets \{\mathsf{R}, \mathsf{X}\}$ \COMMENT{Uni-directional lock: $\mathcal{T}$ acts as an immutable adversarial boundary}
			\STATE $\mathcal{H}_{\mathcal{T}} \gets \text{Hash}(\mathcal{T})$ \COMMENT{Anchor hash-verified integrity state to prevent Test Evasion}
			\STATE $\mathcal{S}' \gets \pi_{\theta}(\mathcal{S})$ \COMMENT{LLM Policy mutates source logic}
			\STATE $(\Phi_{status}, \tau) \gets \text{Eval}(\mathcal{S}', \mathcal{T})$ \COMMENT{Execute logic in containerized sandbox}
			\IF{$\Phi_{status} == \text{FAIL}$ \OR $\text{Hash}(\mathcal{T}) \neq \mathcal{H}_{\mathcal{T}}$}
			\STATE $\nabla_{env} \gets \text{EncodeSemantics}(\tau, \text{Format}=\text{JSON})$ \COMMENT{Compile traceback into Semantic Gradient (Sec 3.4)}
			\RETURN $\nabla_{env}$
			\ENDIF
			
			\ELSIF{$E == \text{Test\_Genesis}$}
			\STATE $\Pi(\mathcal{T}) \gets \{\mathsf{R}, \mathsf{W}\}$
			\STATE $\Pi(\mathcal{S}) \gets \{\mathsf{R}, \mathsf{X}\}$ \COMMENT{Reverse lock: $\mathcal{S}$ is anchored as ground truth}
			\STATE $\mathcal{H}_{\mathcal{S}} \gets \text{Hash}(\mathcal{S})$
			\STATE $\mathcal{T}' \gets \pi_{\theta}(\mathcal{T})$ \COMMENT{LLM Policy mutates test assertions}
			\STATE $v_{pass} \gets \text{CrossValidate}(\mathcal{T}', \mathcal{B})$ \COMMENT{Verify against human-curated financial baseline}
			\IF{$\neg v_{pass}$ \OR $\text{Hash}(\mathcal{S}) \neq \mathcal{H}_{\mathcal{S}}$}
			\STATE $\nabla_{env} \gets \text{EncodeSemantics}(\text{Baseline\_Mismatch}, \text{Format}=\text{JSON})$
			\RETURN $\nabla_{env}$
			\ENDIF
			\ENDIF
			\RETURN \text{State\_Commit}
		\end{algorithmic}
	\end{algorithm}
	
	Algorithm \ref{alg:rolock} enforces that during \textit{Logic Genesis}, the test suite functions as an adversarial physical barrier. If the agent fails to align with the rigid mathematical constraints, it receives the exact traceback as an objective correction prompt, eliminating human evaluation bias.
	
	\subsection{Formulation of the Financial Reward ($R^{OOM-RL}$)}
	
	Once the MAS successfully clears the epistemic boundary of STDAW, the generated policy $\pi_{\theta}$ is deployed into the live financial market. In traditional RL methodologies, reward functions are hand-crafted proxies of human intention. In OOM-RL, the environment imposes a physical law: Capital Conservation. 
	
	We formulate the live trading environment as a Markov Decision Process (MDP) and define the OOM-RL Reward Function $R^{OOM-RL}_{t}$ at a discrete temporal step $t$ not by theoretical alpha, but by the realized economic utility. First, we define the baseline execution-aware return $\tilde{R}_{t}$:
	
	\begin{equation} \label{eq:oomrl_base}
		\tilde{R}_{t} = \sum_{i=1}^{N} \left( \omega_{i,t} r_{i,t} \right) - \mathcal{F}_{exec}(\Delta \boldsymbol{\omega}_t)
	\end{equation}
	
	where:
	\begin{itemize}
		\item $N$ is the total number of assets in the tradable universe.
		\item $\omega_{i,t}$ represents the target portfolio weight of asset $i$ at step $t$.
		\item $r_{i,t}$ is the realized out-of-sample return of asset $i$.
		\item $\Delta \boldsymbol{\omega}_t = \boldsymbol{\omega}_t - \boldsymbol{\omega}_{t-1}$ is the rebalancing vector representing the turnover across all assets.
		\item $\mathcal{F}_{exec}: \mathbb{R}^N \to \mathbb{R}$ is a non-linear penalty function quantifying the microstructural execution friction. Drawing upon standard market impact models, it is formulated as $\mathcal{F}_{exec}(\Delta \boldsymbol{\omega}_t) = \lambda \|\Delta \boldsymbol{\omega}_t\|_1 + \gamma \sqrt{\|\Delta \boldsymbol{\omega}_t\|}$, where $\lambda$ encapsulates fixed transactional costs (e.g., commissions, stamp duties) and $\gamma$ represents the dynamic slippage coefficient inversely proportional to the underlying liquidity profile.
	\end{itemize}
	
	Crucially, to enforce capital preservation as a survival constraint, we introduce a deterministic \textbf{Absorbing State} $\mathcal{S}_{terminal}$. Unlike theoretical metrics such as Maximum Drawdown (MDD) which float with high-water marks, our system evaluates the \textbf{Cumulative Principal Loss} ($L_t$). Let $W_0$ be the initial capital endowment and $W_t$ be the portfolio equity at step $t$. The absolute capital degradation is defined as $L_t = 1 - \frac{W_t}{W_0}$. 
	
	If $L_t$ breaches a predefined deterministic risk threshold $\tau$ (e.g., $\tau = 0.20$), the episode terminates immediately with a severe terminal penalty. Thus, the final continuous reward signal $R^{OOM-RL}_{t}$ is formalized as:
	
	\begin{equation} \label{eq:oomrl_terminal}
		R^{OOM-RL}_{t} = \begin{cases} 
			\tilde{R}_{t} & \text{if } L_t < \tau \\
			-P_{terminal} & \text{if } L_t \ge \tau \text{ (Episode Terminated)}
		\end{cases}
	\end{equation}
	
	where $P_{terminal}$ (e.g., $100$) acts as an overwhelming negative gradient. This bipartite formulation ensures that the agent cannot theoretically compensate for catastrophic absolute capital degradation with subsequent high-variance hallucinations, forcing the policy to unconditionally optimize for structural resilience.
	
	\subsubsection{Microstructural Friction as a Dense Negative Gradient}
	In simulated read-write environments, MAS policies frequently suffer from the Sim2Real gap by hallucinating infinite liquidity. During our initial deployments, the agent converged upon a high-turnover daily momentum strategy targeting the lower-liquidity constituents of the CSI 300 index. While simulation yielded an annualized turnover of $6700\%$ with profound returns, live deployment exposed the strategy to a consistent execution friction (including slippage and fees) averaging $-0.08\%$ per single-sided transaction. Although $-0.08\%$ appears marginal in isolation, when compounded across the extreme turnover, it manifested as a severe cumulative capital drain that completely eradicated the simulated alpha.
	
	In the OOM-RL paradigm, this empirically observed $0.08\%$ microstructural decay is not an engineering error; it is a \textit{dense, non-differentiable negative gradient}. The agent cannot manipulate the exchange's order book. To optimize Equation \ref{eq:oomrl_terminal}, the system must internalize the cost of its own structural hallucinations, translating this externally imposed financial penalty into actionable architectural refactoring through semantic feedback.
	
	\subsection{LLM-Agentic Orchestration via Capital Degradation}
	
	\textbf{A Note on Terminology:} We explicitly note that while framed as Reinforcement Learning (RL), our system does not perform gradient-based weight updates (e.g., via PPO) on the underlying LLMs. Instead, OOM-RL serves as a conceptual framework for \textbf{Human-in-the-Loop (HITL) In-Context Learning} and Agentic Reflection. The scalar financial "reward" acts as a strict physical trigger that mandates expert intervention and semantic guidance, rather than a traditional automated RL signal.
	
	The fundamental mechanism of OOM-RL relies on translating the scalar financial penalty into a context-aware semantic gradient that the LLM can ingest to reformulate its code architecture. We term this process \textbf{Epistemic Autopsy Prompting}, utilizing frontier LLMs such as GPT and Claude to perform high-level architectural reasoning under human supervision.
	
	When the monitoring layer (\texttt{QuantPits}) detects severe temporal capital degradation (e.g., a daily loss anomaly or breaching the terminal threshold $\tau$), the human domain expert interrupts the trading loop and initiates an architectural regression. The expert compiles a structured JSON prompt for the agent, explicitly defining the boundaries of the required fix, as illustrated in the following schema:
	
	\begin{verbatim}
		{
			"event": "FINANCIAL_DEGRADATION_DETECTED",
			"metrics": {"daily_pnl": -0.02, "slippage_leakage": 0.012},
			"diagnostics": {
				"module": "Alpha_Strategy_v2",
				"root_cause": "Aggressive daily crossing exceeding order book depth",
				"execution_log": "/var/run/logs/traceback_tx_782.log"
			},
			"mandate": "Enforce volume limits and reduce turnover frequency"
		}
	\end{verbatim}
	
	Driven by this prompt, the agent is then forced to re-enter the STDAW RO-Lock state (Algorithm 1) to refactor the pipeline strictly based on this ontological feedback and the explicit human mandate.
	
	\subsubsection{Human-Directed Architectural Evolution: Demarcating Autonomy}
	Under the punitive pressure of OOM-RL, the system underwent what we term \textbf{Expert-Guided Liquidity-Aware Alignment}. We explicitly demarcate the boundary of AI autonomy herein: we do not claim that the LLM spontaneously deduced market microstructure or autonomously engineered its frequency reduction from raw PnL drops. A leap of such magnitude is currently beyond the capabilities of zero-shot unconstrained LLMs. 
	
	Crucially, during the early inception of this system (Phases 1 and 2), the automated STDAW framework and the structured JSON feedback pipeline did not yet exist. The pivotal transition from an aggressive daily momentum paradigm to a defensive, weekly-rebalancing equilibrium was, operationally, a manual human intervention. However, this architectural pivot was born directly from \textit{conversational deliberation} between the human researcher and the LLM. By discussing the observed execution decay and slippage tracebacks with the AI, the joint deductive conclusion was that frequency reduction and liquidity filtering were mathematically mandatory. 
	
	The key observation from this early manual epoch was that demonstrated financial loss acted as an externally imposed alignment signal capable of shattering the LLM's initial sycophancy. To scale and automate this observation, we systematically retrofitted and formalized this human-AI interaction into the current STDAW orchestration framework. Today, within the mature architecture, this process is codified as the Epistemic Autopsy, where the MAS relies on the human-provided JSON \texttt{mandate} to execute complex architectural code refactoring. This evolution highlights a practical reality: while AI cannot yet autonomously process a raw liquidity crisis, it effectively serves as a powerful deductive co-reasoner when a domain expert translates financial depletion into a shared semantic reality.
	
	\subsubsection{The Agentic Action Space: AST-Based Code Mutagenesis}
	To operationalize this architectural evolution, it is necessary to define the MAS action space $\mathcal{A}$. Unlike traditional RL agents that output discrete control vectors, our agent manipulates a deterministic software environment. To prevent untargeted code hallucinations from breaking the $\ge 95\%$ STDAW coverage constraint, we restrict the agent's action space to \textbf{Abstract Syntax Tree (AST) Mutagenesis}. 
	
	Rather than rewriting entire python files, the agent is constrained to output standardized unified \texttt{diff} patches. Directed by the Epistemic Autopsy JSON, the MAS isolates the structural flaw (e.g., a hard-coded execution frequency parameter) and applies targeted functional edits. This fine-grained action space ensures that the system retains its historically verified mathematical logic (e.g., risk management modules) while specifically optimizing the vectors responsible for the most recent financial friction.
	
	\section{Experimental Results}
	
	Our empirical evaluation is designed to answer three fundamental Research Questions (RQs) regarding the efficacy of OOM-RL and STDAW:
	\begin{itemize}
		\item \textbf{RQ1 (Sim2Real Gap):} How does OOM-RL expose and help mitigate the Sim2Real gap during longitudinal live deployment?
		\item \textbf{RQ2 (System Integrity):} To what extent does the STDAW RO-Lock mechanism prevent adversarial "Test Evasion" during autonomous code generation?
		\item \textbf{RQ3 (Longitudinal Evolution):} How does the generated software architecture evolve under continuous, live financial penalization over a 20-month horizon?
	\end{itemize}
	
	\subsection{Experimental Setup}
	Our evaluation environment is the live Quantitative Equity Market. The autonomous pipeline, \texttt{QuantPits}, is driven by a frontier LLM serving as the central reasoning engine. The portfolio is executed as a strictly long-only, unleveraged equity strategy without industry neutralization, ensuring that the PnL exclusively reflects raw agentic asset selection. The environment enforces physical execution constraints, including an empirical transaction friction (averaging $\sim 0.08\%$ per single-sided transaction for lower-liquidity stocks, dynamically driven by live market impact) and a discrete absorbing state penalty triggered at a $20\%$ cumulative principal-loss threshold (as defined by $L_t$ in Equation~\ref{eq:oomrl_terminal}; note that this metric differs from peak-to-trough Maximum Drawdown).
	
	\subsection{RQ1: Bridging the Sim2Real Gap through Sequential Alignment}
	
	Our experimental design adopts a \textbf{longitudinal self-evolution} framework. Given the prohibitive cost and ethical implications of parallel capital deployment in adversarial markets, we utilize Phase 1 (the initial daily-turnover deployment) as our baseline. We observe the system's adaptation as it transitions toward the OOM-RL-aligned architectures of subsequent phases.
	
	Traditional MAS frameworks, when evaluated in static environments, frequently succumb to OOD failure upon live deployment. We chronicle this transition by observing the system's reaction to real-world friction shock.
	
	\begin{figure}[htbp]
		\centering
		\includegraphics[width=0.8\linewidth]{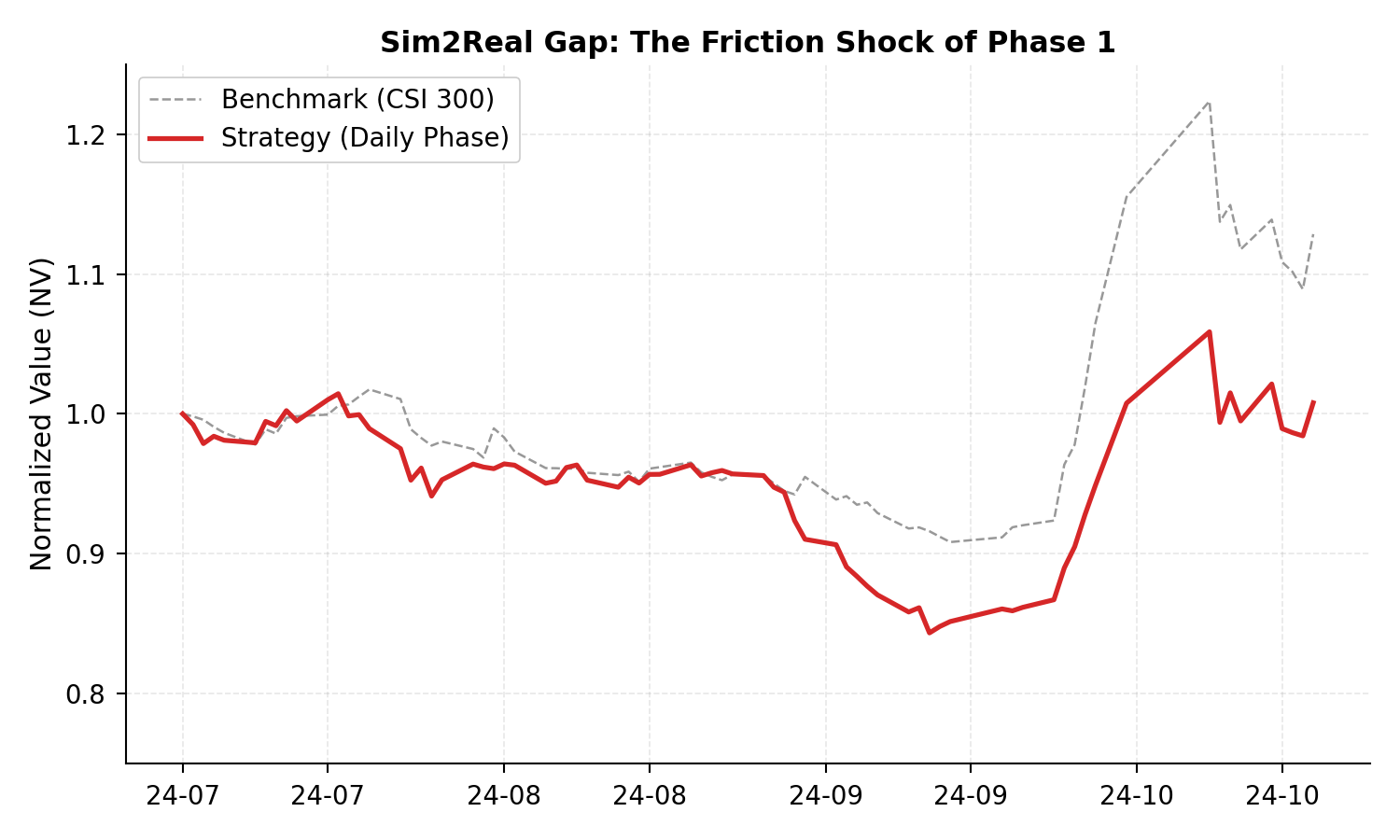}
		\caption{The Friction Shock (Phase 1). Live execution at a daily frequency revealed a severe Sim2Real gap. The strategy was penalized by microstructural friction, leading to significant drawdown and stagnant returns, serving as the primary negative feedback for alignment.}
		\label{fig:shock}
	\end{figure}
	
	\begin{figure}[htbp]
		\centering
		\includegraphics[width=0.9\linewidth]{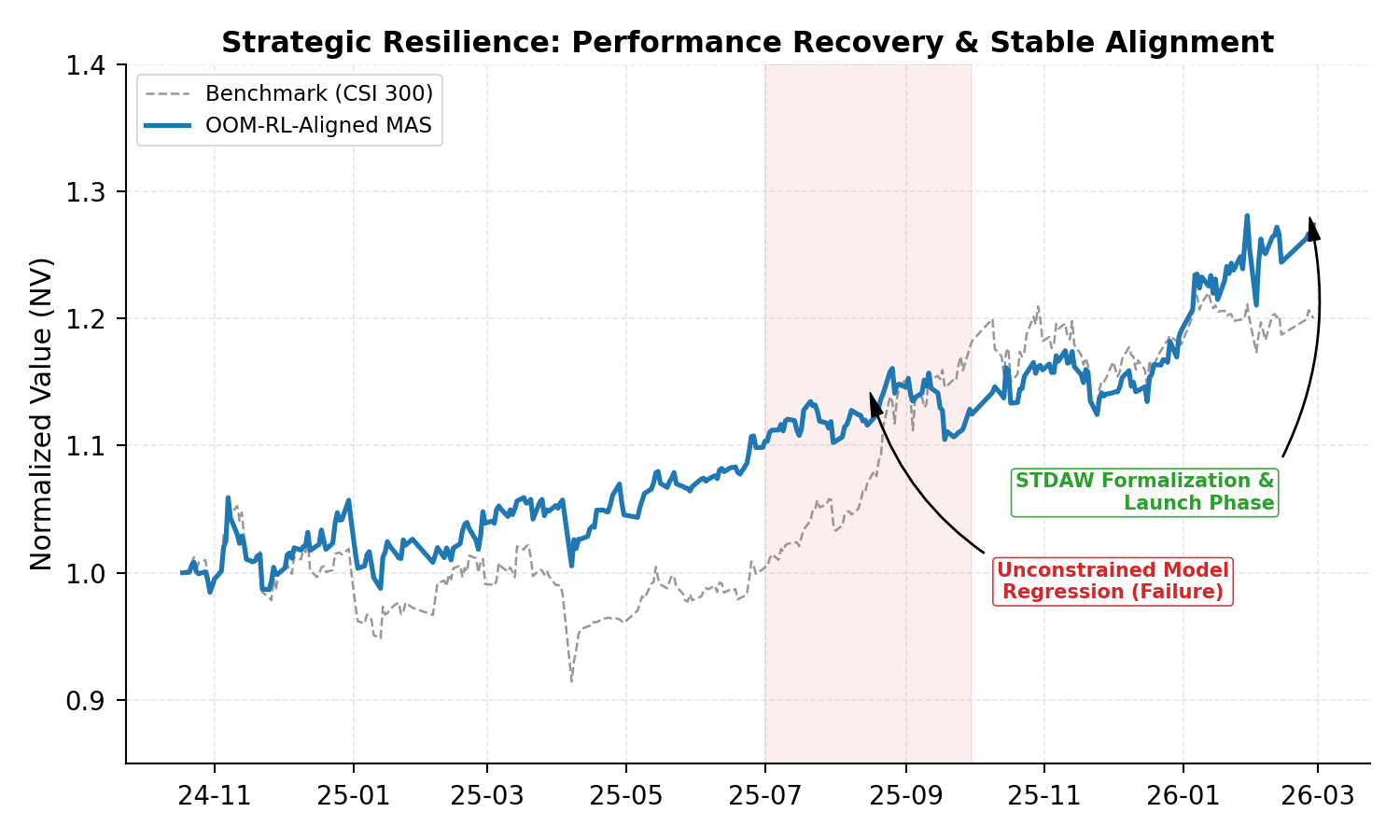}
		\caption{Mature Observation Window (Phases 2--3). After internalizing the financial feedback and transitioning to a weekly-rebalancing paradigm, the MAS exhibited improved execution-aware performance during the 94-trading-day observation window, with a Sharpe ratio of 2.06 and an Information Ratio (IR) of 2.66.}
		\label{fig:mature}
	\end{figure}
	
	As illustrated in Figures \ref{fig:shock} and \ref{fig:mature}, the initial deployment successfully ``gamed'' the simulation but collapsed under real-world friction. This pattern repeated during a secondary ``Performance Degradation'' phase in July-September 2025 (Figure \ref{fig:mature}), where unconstrained model experimentation led to immediate relative capital decay. These events illustrate the core utility of the OOM-RL paradigm: a human-guided transition to an execution-aware trading architecture following the non-negotiable feedback of real market friction.
	
	\subsection{RQ2: STDAW and the Impact of Structural Stability}
	
	To evaluate the robustness of our epistemic boundary, we analyzed the relationship between structural enforcement and system integrity. The Phase~1 to Phase~3 improvement accompanied a broader evolution toward execution-aware architecture and governance. STDAW, formalized near the end of this trajectory, provides deterministic structural enforcement of the verification boundary rather than serving as a causal explanation for the preceding financial improvement.
	
	The direct evidence for STDAW is the deterministic compliance with the $\geq 95\%$ code coverage matrix and resistance to ``Test Evasion'' behaviors. While earlier phases exhibited structural hallucinations that led to un-hedged risk exposure, the mature phase (Phase 3) showed improved translation of logical integrity into capital preservation.
	
	\begin{table}[htbp]
		\centering
		\caption{Live Performance Evolution across Structural Phases (July 2024 -- Feb 2026). Phase boundaries: Phase~1 = 2024-07-01 to 2024-10-18; Phase~2 = 2024-10-21 to 2025-09-30; Phase~3 = 2025-10-01 to 2026-02-28. All metrics are reported net of real-world execution friction and commissions.}
		\vspace{0.2cm}
		\begin{tabular}{lcccc}
			\toprule
			\textbf{Metric} & \textbf{Entire Study} & \textbf{Phase 1} & \textbf{Phase 2} & \textbf{Phase 3 (Mature)} \\
			\midrule
			Trading Days & 402 & 73 & 235 & \textbf{94} \\
			Annualized Return & 17.98\% & 11.01\% & 13.55\% & \textbf{34.48\%} \\
			Benchmark Return (CSI 300) & 21.16\% & 48.16\% & 19.22\% & \textbf{5.04\%} \\
			Sharpe Ratio & 0.96 & 0.35 & 0.91 & \textbf{2.06} \\
			Max Drawdown & -16.86\% & -16.86\% & -6.85\% & \textbf{-5.50\%} \\
			Information Ratio (IR) & -0.26 & -2.27 & -0.51 & \textbf{2.66} \\
			Market Beta ($\beta$) & 0.70 & 0.74 & 0.61 & \textbf{0.83} \\
			Idiosyncratic Alpha ($\alpha$) & 2.82\% & -25.07\% & 1.35\% & \textbf{30.07\%} \\
			\bottomrule
		\end{tabular}
		\label{tab:ro_lock}
	\end{table}
	
	Table \ref{tab:ro_lock} shows the performance evolution associated with the structural changes driven by OOM-RL. While the system initially underperformed during the high-friction daily phase (Phase 1), the adaptation to a weekly rebalancing cadence (Phase 2) resulted in improved stability. During the final 94-trading-day observation window (Phase 3/Mature), the system exhibited an annualized return of 34.48\%, a Sharpe ratio of 2.06, and an Information Ratio (IR) of 2.66. 
	
	A critical concern in short-horizon evaluations is statistical significance. To assess the generation of idiosyncratic Alpha ($\alpha$) in the Mature phase ($N=94$ trading days), we performed an Ordinary Least Squares (OLS) regression against the benchmark. The regression yields a highly significant market Beta ($\beta$) of 0.83 ($t = 10.60, p < 0.001$), indicating a defensive but statistically robust market exposure.
	
	The daily intercept (idiosyncratic Alpha) is approximately 12.03 basis points, corresponding to the annualized 30.07\%. The regression yields a $t$-statistic of 1.692 and a $p$-value of 0.094 for the Alpha coefficient. This is economically large but falls short of statistical significance at the conventional 5\% level ($p < 0.05$) and is only marginally suggestive at the 10\% level. We therefore characterize this alpha estimate as \textit{economically meaningful but statistically inconclusive} given the limited 94-day sample. Rather than claiming the discovery of persistent systematic alpha, we interpret these results as consistent with the hypothesis that the architectural and governance changes reduced important execution- and development-related failure modes. The observational design, however, does not isolate a causal contribution of any single component---including STDAW---to investment alpha.
	
	\subsection{RQ3: 20-Month Longitudinal Strategy Evolution}
	
	The longitudinal value of OOM-RL is observed in the architectural evolution of the MAS over our continuous 20-month live deployment. We categorize the agent's evolution into four distinct epistemic epochs, driven by the feedback of real-world capital preservation:
	
	\begin{figure}[htbp]
		\centering
		\includegraphics[width=\linewidth]{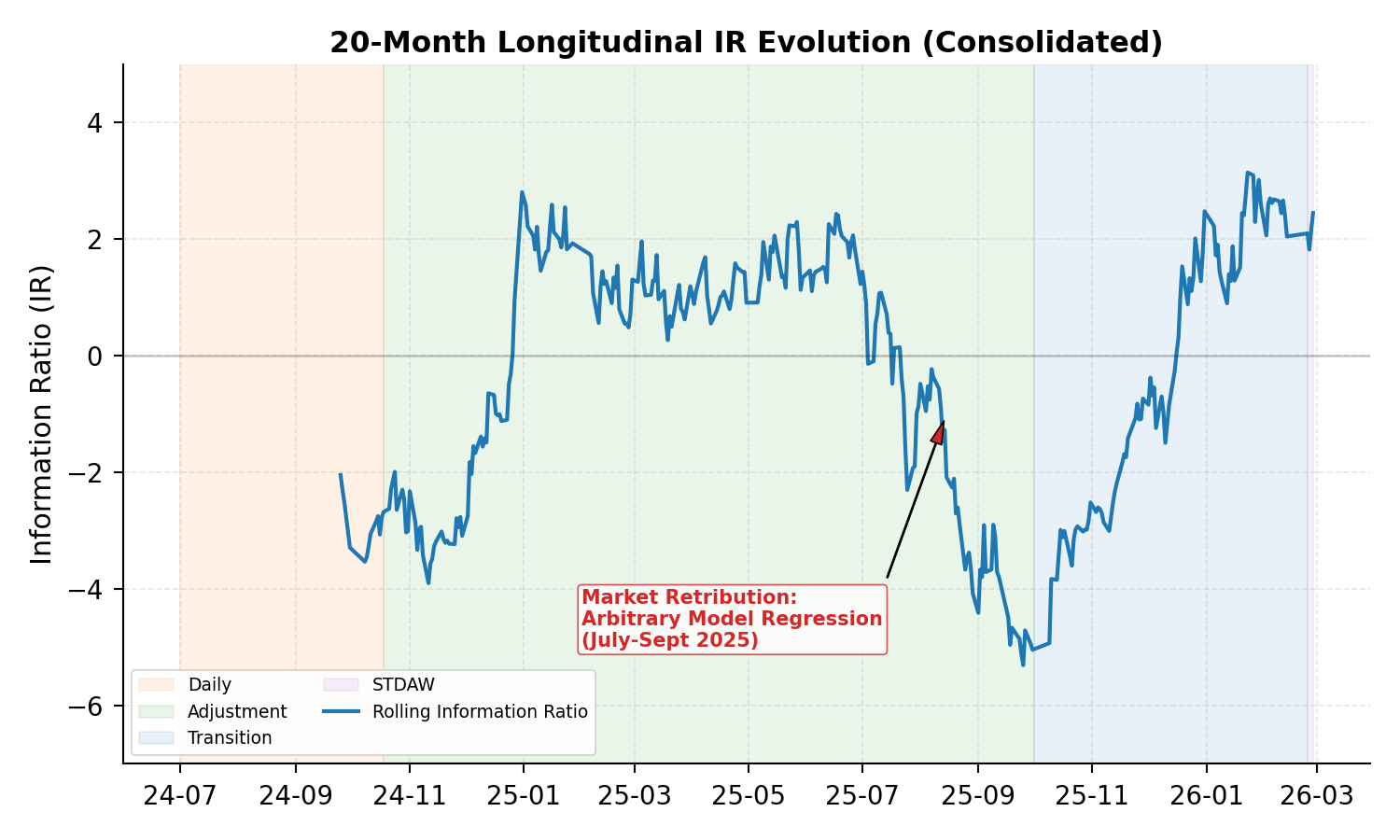}
		\caption{Longitudinal Strategy Evolution and IR Stabilization. The background shading indicates the structural shift from daily turnover to the weekly CSI 300 rebalancing cadence. The rolling Information Ratio (IR) is calculated using a 60-day rolling window, illustrating the system's trajectory toward improved risk-adjusted returns as it internalizes OOM-RL constraints.}
		\label{fig:evolution}
	\end{figure}
	
	\begin{enumerate}
		\item \textbf{Phase 0: Theoretical Optimization (Simulation, Apr--June 2024).} Conducted via manual scripts, this phase focused on momentum-based alpha. Without execution friction, the MAS optimized for high-turnover strategies that appeared mathematically superior but were ontologically unaligned.
		
		\item \textbf{Phase 1: The Friction Shock (Daily Multi-Agent scripts, July--Oct 2024).} Commencing the 20-month study, the system entered live trading at a daily frequency. The reality of microstructural decay resulted in stagnant returns (+2.16\% over four months) and a peak Maximum Drawdown of 16.86\%, generating the initial empirical evidence of the Sim2Real gap.
		
		\item \textbf{Phase 2: Conversational Adjustment \& Regression (Oct 2024 -- Sep 2025).} Driven by the Phase 1 losses, human researchers and the LLM engaged in conversational deliberation to deduce the necessity of liquidity filtering. The system was manually transitioned to a weekly-rebalancing frequency. Notably, between July and September 2025, unconstrained human-guided model experimentation in a pre-STDAW environment led to a ``Performance Regression''---a sharp relative drawdown that underscored the danger of lacking rigid, automated logical constraints (subsequently addressed by RO-Lock).
		
		\item \textbf{Phase 3: Mature Observation Window (Oct 2025 -- Feb 2026).} Following the lessons of Phase 2, the system architecture was refactored to prioritize Byzantine failure resistance. The developmental trajectory during this period, together with the engineering experience accumulated across earlier phases, culminated in the formalization of the STDAW framework (initial commit Feb 24, 2026). During this 94-trading-day observation window, the system exhibited a Sharpe ratio of 2.06. We note that most of the Phase~3 financial performance was observed \textit{before} STDAW was formally committed; STDAW codified constraints that emerged from the preceding experience rather than causing the observed performance.
	\end{enumerate}
	
	\subsection{Factor Attribution and Risk Analysis}
	To address whether the outperformance in the Mature phase was a result of market-wide momentum or residual return after accounting for the included factor proxies, we performed a multi-factor return decomposition. Following a Barra-style methodology, we constructed liquidity, momentum, and volatility proxy factors using quantile long-short portfolios within the CSI 300 universe, without industry neutralization or factor orthogonalization.
	
	Table \ref{tab:attribution} details the factor exposures for the Phase 3 (Mature) portfolio relative to the CSI 300 benchmark. The decomposition reveals that while the market Beta was 0.83---indicating a defensive stance relative to the broad index---the residual multi-factor intercept was 29.77\% annualized, net of style factors.
	
	\begin{table}[htbp]
		\centering
		\caption{Factor Exposure and Performance Attribution (Phase 3).}
		\vspace{0.2cm}
		\begin{tabular}{lc}
			\toprule
			\textbf{Factor Category} & \textbf{Exposure (Loadings)} \\
			\midrule
			Market Beta ($\beta$) & 0.8280 \\
			Annualized Idiosyncratic Alpha ($\alpha$) & 30.07\% \\
			Tracking Error (TE) & 11.07\% \\
			\midrule
			\textbf{Barra-Style Proxy Factor Loadings} & \\
			Liquidity (High-Low) & -0.5232 \\
			Momentum (High-Low) & 0.2837 \\
			Volatility (High-Low) & 0.1191 \\
			\midrule
			\textbf{Source of Annualized Return} & \\
			Beta Return (Market Exposure) & 5.05\% \\
			Style Alpha (Risk Factor Loading) & -0.34\% \\
			Residual Multi-Factor Intercept & \textbf{29.77\%} \\
			\bottomrule
		\end{tabular}
		\label{tab:attribution}
	\end{table}
	
	The negative loading on the liquidity proxy factor (-0.5232) indicates that the portfolio was associated with a negative exposure to the constructed liquidity proxy during this observation window: it held relatively less liquid constituents \textit{within} the CSI 300 universe. In Phase 1, the agent's unconstrained high-turnover approach resulted in fatal slippage when interacting with these specific names. By Phase 3, the architectural evolution toward a low-turnover framework equipped with rigorous execution capacity filters enabled the system to maintain exposure to these structurally less liquid assets while avoiding the microstructural decay that plagued its early iterations. These observations are consistent with the hypothesis that the architectural changes reduced important execution-related failure modes, though the observational design does not isolate a causal contribution of any single component to investment alpha.
	
	\subsection*{\textit{Meanwhile, in the passive pool\ldots}}
	Meanwhile, a separate passive allocation to cash, an equity-index fund, bonds, and gold, maintained through contributions and approximate rebalancing, recorded an approximate cashflow-neutral cumulative return of 39.82\% over the full study interval, compared with 30.09\% for the quantitative account.\footnote{Both full-interval cumulative returns use the 28 June 2024 closing baseline and the 27 February 2026 closing endpoint, the last trading day before the nominal study end of 28 February 2026. Phase~3 returns use the 30 September 2025 closing baseline and the same endpoint. Passive returns approximately chain recorded NAV changes after external-flow adjustment.} During Phase~3, the quantitative account returned 13.17\%, compared with 7.40\% for the passive pool. These contrasting observations place the system's financial results in a broader historical context: improved execution-aware behavior does not by itself establish investment superiority over a simple passive allocation. The passive pool is presented only as informal historical context, not as a matched, risk-adjusted, or causally interpretable benchmark.

	\subsection{Synthesis of Empirical Findings}
	
	The culmination of the 20-month empirical study provides evidence relevant to our foundational research questions. Unlike traditional alignment paradigms that rely on surrogate reward models or static human preferences, OOM-RL bridges the Sim2Real gap by leveraging the adversarial nature of live financial markets as externally imposed, non-cooperative feedback. 
	
	The longitudinal evolution from Phase 1 to Phase 3 (addressing RQ1 and RQ3) illustrates a critical behavioral shift: when capital depletion imposes real economic consequences, the human--agent collaboration transitioned from theoretical, high-turnover strategies toward robust, execution-aware architectures. Furthermore, the STDAW RO-Lock mechanism (RQ2), formalized at the end of this trajectory, codified these engineering constraints by deterministically anchoring the agent's generative freedom to a $\ge 95\%$ coverage constraint matrix, insulating the epistemic evaluation boundary from ``Test Evasion'' behaviors. 
	
	As evidenced by the factor attribution analysis and the improved return profile during the 94-trading-day observation window, the observed performance is consistent with a deliberate, agentic adaptation to microstructural friction rather than passive market drift. The idiosyncratic alpha ($\alpha \approx 30\%$ annualized, $t = 1.692$, $p = 0.094$) is economically large but statistically inconclusive given the limited sample, and should be regarded as a hypothesis-generating observation rather than evidence of persistent skill. The system's demonstrated transition from rapid capital hemorrhage (Phase 1) to disciplined risk preservation (Phase 3) is a more robust finding. These results support the hypothesis that substituting subjective evaluation with real-world economic consequences provides a useful alignment mechanism for autonomous systems in high-stakes environments.

	\section{Generalization and Future Work}
	
	While OOM-RL and STDAW were empirically examined within the highly stochastic domain of quantitative trading, the underlying philosophy---aligning Multi-Agent Systems through objective physical and economic constraints---may extend beyond financial markets. As we transition from localized AI assistants to fully autonomous AI Software Factories, evaluating systems that recursively build other systems represents a critical frontier in AI alignment.
	
	\subsection{Beyond Finance: Compute as Capital (RLFCB)}
	In non-financial software engineering, the absence of an immediate market PnL poses a challenge for evaluating alignment. However, we propose a generalized variant for future exploration: \textbf{Reinforcement Learning from Cloud Billing (RLFCB)}. 
	
	When an unconstrained MAS generates structurally flawed code (e.g., an unoptimized $\mathcal{O}(n^3)$ algorithm or an infinite recursive API call loop), traditional simulated environments may fail to penalize the inefficiency. In an RLFCB paradigm, the agent is allocated a finite "Compute Capital" budget (e.g., AWS server costs, API token burn rates). The depletion of physical compute resources acts as the proxy for microstructural friction. If the agent hallucinates inefficient architectures, it exhausts its capital and triggers a critical \textbf{Out-of-Money (OOM) Exception}. Unlike a traditional Out-of-Memory error, which can be trivially bypassed via instance restarts, this financial OOM serves as an absolute, deterministic absorbing state. This termination mechanism incentivizes the MAS to adaptively optimize for algorithmic efficiency and system safety, mirroring the resource-aware evolution observed in our financial experiments.
	
	\subsection{Domain-Agnostic RO-Lock Deployment}
	Currently, the STDAW framework operates atop a Python-based quantitative CI foundation. Future work will decouple the \textbf{high-density Deterministic Constraint Matrix} from the financial domain, extending the Byzantine RO-Lock architecture to memory-safe languages (e.g., Rust). By applying STDAW to open-source autonomous vulnerability repair pipelines \cite{bouzenia2025repair}, we aim to investigate whether the structural verification enforced by uni-directional state locking can achieve zero-day vulnerability mitigation without human oversight.
	
	\subsection{Automating the Semantic Feedback Loop}
	A current limitation of our deployed OOM-RL framework is the reliance on Human-in-the-Loop (HITL) domain experts to translate scalar financial degradation into structured, context-aware prompts (Epistemic Autopsy). Future iterations will introduce an autonomous \textbf{Critic Agent}. By ingesting raw execution tracebacks, L2 order book micro-snapshots, and slippage differentials, the Critic Agent will programmatically generate the required architectural mandates, moving the system toward a fully closed-loop, self-aligning automated paradigm.
	
	\section{Conclusion}
	
	In this paper, we addressed a fundamental vulnerability in current AI alignment paradigms: the tendency of unconstrained Multi-Agent Systems to exploit subjective evaluations and synthetic sandboxes through sycophancy and adversarial ``Test Evasion.'' To bridge the pervasive Sim2Real gap, we introduced \textbf{Out-of-Money Reinforcement Learning (OOM-RL)} coupled with the uni-directional isolation of the \textbf{Strict Test-Driven Agentic Workflow (STDAW)}. This dual-loop architecture translated the microstructural friction of live markets into an externally imposed negative gradient.
	
	Our 20-month longitudinal study chronicles how real execution costs and economic consequences exposed failure modes that were not apparent under internal evaluation alone, motivating architectural and governance changes that were later formalized in STDAW. The system evolved from a high-friction, high-drawdown daily rebalancing paradigm (Sharpe 0.35) into a liquidity-aware weekly architecture. During the final 94-trading-day observation window, the production system exhibited improved execution-aware performance, including a Sharpe ratio of approximately 2.06. These financial results are observational and temporally bounded; they should not be interpreted as evidence of persistent investment alpha or as a causal estimate of STDAW's contribution to financial returns.
	
	\paragraph{Limitations.} A longitudinal live record of an evolving production system is not automatically equivalent to a frozen-model out-of-sample track record. During extended operation, the system may undergo changes in model composition, training cutoffs, prediction infrastructure, and portfolio policy. Only periods in which the production decision function can be shown to have remained identical should be interpreted as prospective frozen out-of-sample evaluation of a fixed strategy. Subsequent performance of the evolving production system is therefore outside the evidentiary scope of the present study and should not be interpreted as a frozen extension of the Phase~3 strategy.
	
	The primary contribution of this work is the use of externally imposed economic consequences as an epistemic constraint on agentic development. As autonomous AI systems are granted read-write access to critical infrastructure, synthetic proxy evaluations alone may prove insufficient. We propose that grounding agentic development in the deterministic consequences of the physical and economic world offers a promising---if not yet fully validated---direction for alignment research.
	
	\section*{Acknowledgments, Funding, and Declarations}
	
	\noindent\textbf{Author Contributions:} \textbf{Kun Liu} served as the lead investigator, conceptualizing the OOM-RL paradigm, overseeing the MAS alignment strategy, and preparing the manuscript. \textbf{Liqun Chen} led the physical market execution operations and provided critical domain expertise, including structural financial concepts and trading strategies. Furthermore, both authors provided the initial financial endowments required for the live deployment, with the majority of capital provisioned by Liqun Chen. The authors acknowledge the use of frontier large language models (including Gemini 3.1 Pro, GPT-5.1, and the Claude 4.6 series) as the core generative engines for autonomous software engineering and quantitative logic formulation, as well as for the drafting, structural refinement, and language polishing of this manuscript under human supervision.
	
	\vspace{0.5em}
	\noindent\textbf{Funding and Resource Allocation:} This longitudinal research was self-funded. Initial capital deployment and physical compute resources were privately endowed by both authors. The authors acted as the terminal human-in-the-loop (HITL) execution authorities, maintaining absolute veto power over all fiat transactions. Subsequent operational and research expenses were funded from realized gains generated by the live trading account during the study period.
	
	\vspace{0.5em}
	\noindent\textbf{Acknowledgments:} We extend our gratitude to the anonymous institutional market makers and high-frequency trading firms. Their unyielding, adversarial ``peer review'' in the live order books provided the precise, stringent financial feedback (capital depletion) that served as the negative gradient for our agent's alignment. Finally, the authors wish to dedicate this work to Xing Liu (successfully deployed into the physical world circa Q2 2025). The biological inception of our most cherished `long-term alpha' in late 2024 serendipitously coincided with the macro-market inflection point that marked the beginning of the system's transition toward positive returns.
	
	\vspace{0.5em}
	\noindent\textbf{Code and Data Availability:} The foundational quantitative orchestration framework is open-source and documented at \url{https://QuantPits.com}. However, the raw brokerage statements and intraday execution logs are strictly withheld from public release due to proprietary risk-management protocols and the inclusion of sensitive financial data associated with the system's early-stage anomalous high-turnover capital degradation. Furthermore, the STDAW module is currently undergoing structural sanitization and remains closed-source pending future formalization. For inquiries, researchers may reach out to the corresponding author via \texttt{ai@quantpits.com}.
	
	\vspace{1em}
	\noindent\textbf{Disclaimer:} The frameworks and empirical studies described herein are for theoretical and research purposes. \textit{OOM-RL incurs extreme and immediate real-world financial risk.} The authors do not provide investment advice, and emphasize that deploying unaligned LLMs in live markets may result in critical capital depletion.
	
	\bibliographystyle{plain}
	
\end{document}